\pdfoutput=1
\documentclass{article}

    \PassOptionsToPackage{numbers,compress}{natbib}

     \usepackage[preprint]{neurips_2019}

\usepackage{lipsum}
\usepackage[utf8]{inputenc} 
\usepackage[T1]{fontenc}    
\usepackage{hyperref}       
\usepackage{url}            
\usepackage{booktabs}       
\usepackage{amsfonts}       
\usepackage{nicefrac}       
\usepackage{microtype}      
\usepackage{graphicx}
\usepackage{wrapfig}
\usepackage{amsmath}

\DeclareMathOperator*{\argmin}{arg\,min}
\newcommand\blfootnote[1]{%
  \begingroup
  \renewcommand\thefootnote{}\footnote{#1}%
  \addtocounter{footnote}{-1}%
  \endgroup
}
\title{Not All Features Are Equal: Feature Leveling Deep Neural Networks for Better Interpretation}

\author{%
  Yingjing Lu* \\
  Carnegie Mellon University\\
  \texttt{yingjinl@andrew.cmu.edu} \\
\And
   Runde Yang* \\
   Cornell University \\
   \texttt{ry82@cornell.edu} \\
}

\begin{document}

\maketitle

\begin{abstract}
  Self-explaining models are models that reveal decision making parameters in an interpretable manner so that the model reasoning process can be directly understood by human beings. General Linear Models (GLMs) are self-explaining because the model weights directly show how each feature contributes to the output value. However, deep neural networks (DNNs) are in general not self-explaining due to the non-linearity of the activation functions, complex architectures, obscure feature extraction and transformation process. In this work, we illustrate the fact that existing deep architectures are hard to interpret because each hidden layer carries a mix of low level features and high level features. As a solution, we propose a novel feature leveling architecture that isolates low level features from high level features on a per-layer basis to better utilize the GLM layer in the proposed architecture for interpretation. Experimental results show that our modified models are able to achieve competitive results comparing to main-stream architectures on standard datasets while being more self-explainable. Our implementations and configurations are publicly available for reproductions$\dagger$.
\end{abstract}

\section{Introduction}
Deep Neural Networks (DNNs) are viewed as back-box models because of their obscure decision making process. As a result, those models are hard to verify and are susceptible to adversarial attacks. Thus, it is important for researchers to find ways to interpret DNNs to improve their applicability.

One reason that makes deep neural networks hard to interpret is that they are able to magically extract abstract concepts through multi-layer non-linear activations and end-to-end training. From a human perspective, it is hard to understand how features are extracted from different hidden layers and what features are used for final decision making. 

In response to the challenge of interpretability, two paths are taken to unbox neural networks' decision learning process. One method is to design verifying algorithms that can be applied to existing models to back-trace their decision learning process. Another method is to design models that "explain" the decision making process automatically. The second direction is promising in that the interpretability is built-in architecturally. Thus, the verification feedback can be directly used to improve the model.
\blfootnote{$\dagger$ Public Repo URL annonymized for review purpose-See supplementals for detailed implementation}

One class of the self-explaining models borrows the interpretability of General Linear Models (GLMs) such as linear regression. GLMs are naturally interpretable in that complicated interactions of non-linear activations are not involved. The contribution of each feature to the final decision output can simply be analyzed by examining the corresponding weight parameters. Therefore, we take a step forward to investigate ways to make DNNs as similar to GLMs as possible for interpretability purpose while maintaining competitive performance. 

Fortunately, a GLM model naturally exists in the last layer of most common architectures of DNNs (See supplemental for the reason that the last layer is a GLM layer). However, the GLM could only account for the output generated by the last layer and this output is not easy to interpret because it potentially contains mixed levels of features. In the following section, we use empirical results to demonstrate this mixture effect. Based on this observation, one way to naturally improve interpretation is to prevent features extracted by different layers from mixing together. Thus, we directly pass features extracted by each layer to the final GLM layer. This can further improve interpretability by leveraging the weights of the GLM layer to explain the decision making process. Motivated by this observation, we design a feature leveling network structure that can automatically separate low level features from high level features to avoid mixture effect. In other words, if the low level features extracted by the $k^{th}$ hidden layer can be readily used by the GLM layer, we should directly pass these features to the GLM rather than feeding them to the $k+1^{th}$ hidden layer. We also propose a feature leveling scale to measure the complexity of different sets of features' in an unambiguous manner rather than simply using vague terms such as "low" and "high" to describe these features.

In the following sections, we will first lay out the proposed definition of feature leveling. We then will illustrate how different levels of features reside in the same feature space. Based on the above observations, we propose feature leveling network, an architectural modification on existing models that can isolate low level features from high level features within different layers of the neural network in an unsupervised manner. In the experiment section, we will use empirical results to show that this modification can also be applied to reduce the number of layers in an architecture and thus reduce the complexity of the network. In this paper, we focus primarily on fully connected neural networks(FCNN) with ReLU activation function in the hidden layers. Our main contributions are as follows:
\begin{itemize}
    \item We take a step forward to quantify feature complexity for DNNs.
    \item We investigate the mixture effect between features of different complexities in the hidden layers of DNNs.
    \item We propose a feature leveling architecture that is able to isolate low level features from high level features in each layer to improve interpretation.
    \item We further show that the proposed architecture is able to prune redundant hidden layers to reduce DNNs' complexity with little compromise on performance.  
\end{itemize}

The remaining content is organized as follows: In section 2, we first introduce our definitions of feature leveling and use a toy example to show the mixture effect of features in hidden layers. In section 3, we give a detailed account of our proposed feature leveling network that could effectively isolate different levels of features. In section 4, we provide a high level introduction to some related works that motivated our architectural design. In Section 5, we test and analyze our proposed architecture on various real world datasets and show that our architecture is able to achieve competitive performance while improving interpretability. In section 6, we show that our model is also able to automatically prune redundant hidden layers, thus reducing the complexity of DNNs.

\section{Feature leveling for neural networks}

The concepts of low level and high level features are often brought up within the machine learning literature. However, their definitions are vague and not precise enough for applications. Intuitively, low level features are usually "simple" concepts or patterns whereas high level features are "abstract" or "implicit" features. 

Within the scope of this paper, we take a step forward to give a formal definition of feature leveling that quantizes feature complexity in an absolute scale. This concept of a features' scale is better than simply having "low" and "high" as descriptions because it reveals an unambiguous ordering between different sets of features. We will use a toy example to demonstrate how features can have different levels and explain why separating different levels of features could improve interpretability. 

\subsection{ A toy example}
We create a toy dataset called Independent XOR(IXOR). IXOR consists of a set of uniformally distributed features $\mathcal{X}: \{(x^1, x^2, x^3) | x^1 \in [-2, 2], x^2 \in [-2, 2], x^3 \in [0, 1]\}$ and a set of labels $\mathcal{Y}:\{0, 1\}$. The labels are assigned as:
\[ 
\left \{
\begin{array}{ll}
     &  y = 1 ~~~ x^1 \times x^2 > 0 \wedge x^3 > 0.5 \\
     & y = 0 ~~~ otherwise
\end{array}
\right.
\]

\begin{figure}[h!]
  \centering
  \includegraphics[width=0.5\linewidth]{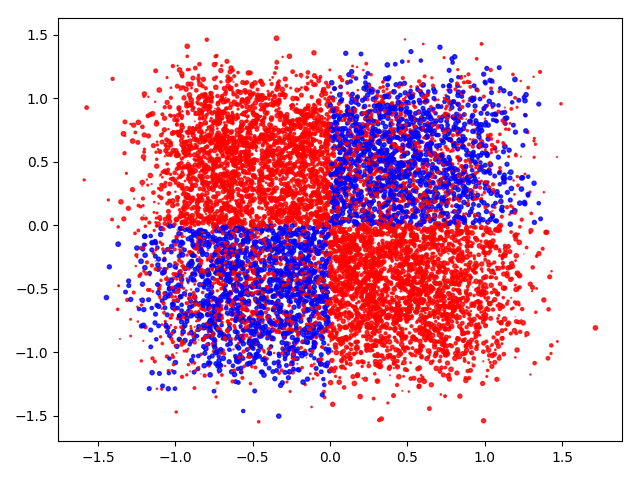}
  \caption{Visualization of the toy IXOR dataset}
\end{figure}

In this dataset, $(x^1, x^2, x^3)$ clearly have different levels of feature. $x^3$ can be directly used by the GLM layer as it has a linear decision boundary. $(x^1, x^2)$ is more complex as they form an XOR pattern and cannot be linearly separated, thus requiring further decomposition to be made sufficient for the GLM layer. To make correct decisions, the DNN should use one layer to decompose the XOR into lower level features, and directly transport $x^3$'s value to into the GLM layer. 

\subsection{Characterize low and high level features with feature leveling}
From IXOR we can see that not all features have the same level of "complexity". Some could be directly fed into the GLM layer, others may need to go through one or more hidden layers to be transformed to features that can directly contribute to decision making. 

Thus, instead of using "low" and "high" level to characterize features, we propose to frame the complexity of different features with the definition of feature leveling.

For a dataset $\mathcal{D}$ consisting of $N$ i.i.d samples with features and their corresponding labels $\{(x_1, y_1),...,(x_N, y_N)\}$. We assume that samples $x_i \in \mathcal{D}$ contains features that requires at most $K$ hidden layers to be transformed to perform optimal inference.

For a DNN trained with K hidden layers and a GLM layer, we define the set of $k^{th}$ level feature as the set of features that requires $k-1$ hidden layers to extract under the current network setup to be sufficiently utilized by the GLM layer. In the following paragraphs, we denote $l_k \in L_k$ as the $k^{th}$ level features extracted from one sample and $L_k$ denotes the set of all $k^{th}$ level feature to be learned in the target distribution. The rest of high level features are denoted by $h_k$ that should be passed to the $k^{th}$ layer to extract further level features. In this case, $l_k$ and $h_k$ should be disjoi[nt, that is $l_k \bigcap h_k = \emptyset$. 
In the case of the toy example, $x^3$ is $l_1$, level one feature, as it is learned by the first hidden layer to directly transport its value to the GLM layer. $(x^1, x^2)$ is $h_1$. The XOR can be decomposed by one hidden layer with sufficient number of parameters to be directly used by the GLM layer to make accurate decisions. Assuming the first hidden layer $f_1$ has sufficient parameters, it should take in $h_1$ and output $l_2$. 

\subsection{How the proposed model solves the mixture effect and boosts interpretation}
However, common FCNN does not separate each level of feature explicitly. Figure 2 shows the heatmaps of the weight vectors for both FCNN baseline and proposed feature leveling network trained on the IXOR dataset. We observe from FCNN that $x^3$'s value is able to be preserved by the last column of the weight vector from the first layer but is mixed with all other features in the second layer, before passing into the GLM layer. Our proposed model, on the other hand, is able to cleanly separate $x^3$ and preserve its identity as an input to the GLM layer. In addition, our model is able to identify that the interaction between $(x^1, x^2)$ can be captured by one single layer. Thus, the model eliminates the second layer and pass $(x^1, x^2)$ features extracted by the first hidden layer directly to the GLM layer. 

Looking at the results obtained from the toy example, we can clearly see that the proposed model is able to solve the mixture effect of features and gives out correct levels for features with different complexities in the context of the original problem. Therefore, the model is more interpretable in that it creates a clear path of reasoning and the contirbution of each level of features can be understood from the weight parameters in the GLM. 
\begin{figure}[h!]
  \centering
  \includegraphics[width=1.0\linewidth]{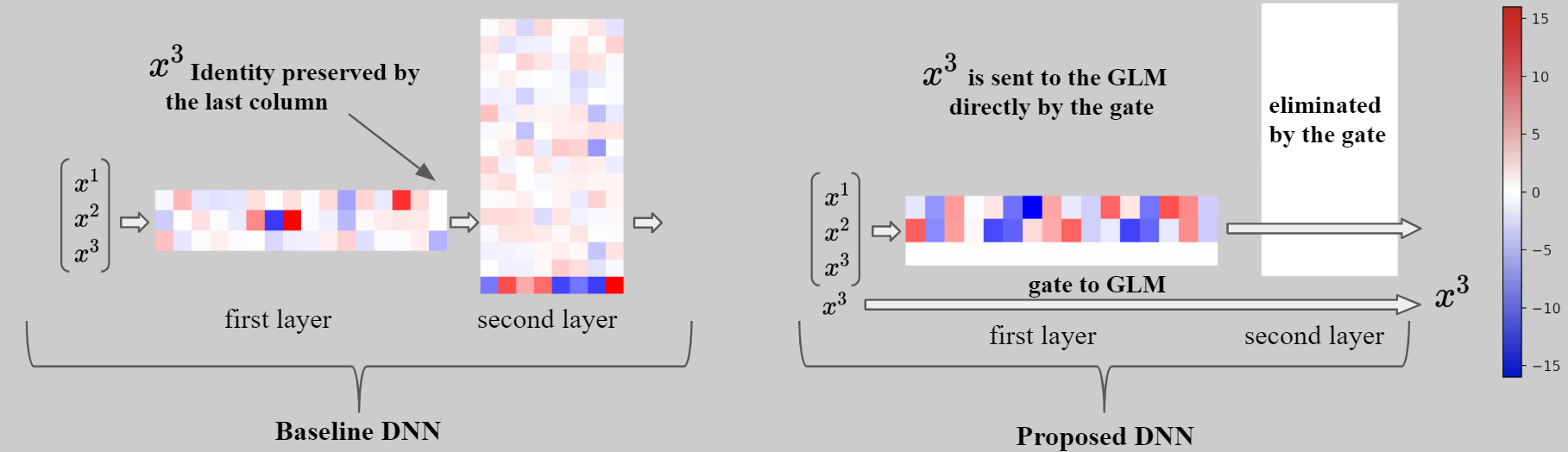}
  \caption{Weight heatmap of Baseline and proposed model with the initial architecture of 3-16-8-2. Arrows denotes information flow. $x^3$ in the proposed model is gated from mixing with other features input to the hidden layer.}
\end{figure}

\section{Our proposed architecture}
Inspired by our definition of feature leveling and to resolve the mixture of features problem, we design an architecture that is able to recursively filter the $k^{th}$ level features from the $k^{th}$ layer inputs and allow them to be directly passed to the final GLM layer. 

We start with a definition of a FCNN and extend that to our model: we aim to learn a function $\mathcal{F}$ parametrized by a neural network with $K$ hidden layers. The function $\mathcal{F}$ can be written as:
\begin{equation}
    \mathcal{F} = d\Big( f_K( f_{K-1}( ... f_1( x; \theta_1 ) ); \theta_K )\Big)
\end{equation}
$f_k$ is the $k^{th}$ hidden layer function with parameters $\theta_k$. $d(\cdot)$ is the GLM model used for either classification, or regression. Thus, the goal is to learn the function $\mathcal{F}$ such that:
\begin{equation}
    \mathcal{R}(\theta) = \frac{1}{N}\bigg( \sum_{i=1}^N  \mathcal{L}(\mathcal{F}(x_i;\theta), y_i)\bigg) ~~~~~~~~\theta^* = \argmin_{\theta}(\mathcal{R}(\theta))
\end{equation}

In our formulation, each hidden layer can be viewed as separator for the $k^{th}$ level features and extractor for higher level features. Thus, the output of $f_k$ has two parts: $l_k$ is the set of $k^{th}$ level feature extracted from inputs and can be readily transported to the GLM layer for decision making. And $h_k$ is the abstract features that require further transformations by $f_k$. In formal language, we can describe our network with the following equation ("$-$"denotes set subtraction):
\begin{equation}
    \mathcal{F} = d\Big( l_1, l_2,...l_{K}, f_K( f_{K-1}( ... f_1( x-l_1 ;\theta_{1} ))-l_{K} ) \Big)
\end{equation}

In order for $f_k$ to learn mutually exclusive separation, we propose a gating system for layer $k$, paramatrized by $\phi_k$, that is responsible for determining whether a certain dimension of the input feature should be in $l_k$ or $h_k$. For a layer with input dimension $J$, the gate $\{z^1_k, ...z^J_k\}$ forms the corresponding gate where $z^j_k \in \{0, 1\}$. $\phi_k$ is the parameter that learns the probability for the gate $z^j_k$ to have value 1 for the input feature at $j^{th}$ dimension to be allocated to $h_k$ and $l_k$ otherwise. 

In order to maintain mutual exclusiveness between $l_k$ and $h_k$, we aim to learn $\phi_k$ such that the it allows a feature to pass to $l_k$ if and only if the gate is exactly zero. Otherwise, the gate is 1 and the feature goes to $h_k$. Thus, we can rewrite the neural network $\mathcal{F}$ with the gating mechanism for the $i^{th}$ sample $x_i$ from the dataset:

\begin{equation}
    \mathcal{F} = d\Big( B( z_1 )\odot x_i, B(z_2)\odot f_1( z_1 \odot x_i ), ..., f_K(z_K \odot f_{K-1}( z_{K-1} \odot f_{K-2}( ...f_1( z_1 \odot x_i ) ) )) ) \Big)
\end{equation}

Here, $\odot$ acts as element-wise multiplication. The function $B$ acts as a binary activation function that returns 1 if and only if the value of $z$ is 0 and 0 otherwise. The function $B$ allows level k feature $l_k = B(z_k) \odot f_{k-1}$ to be filter out if and only if it does not flow into the next layer at all. 

Then the optimization objective becomes:
\begin{equation}
    \mathcal{R}(\theta, \phi) = \frac{1}{N}\bigg( \sum_{i=1}^N ( \mathcal{L}(\mathcal{F}(x_i, z;\theta, \phi, B), y_i)\bigg) + \lambda \sum_{k=1}^K ||z_k||_0 ~,~~~ z_k = g(\phi_k)
\end{equation}
With an additional $L_0$ regularization term to encourage less $h_k$ to pass into the next layer but more $l_k$ to flow directly to the GLM layer. $g(\phi)$ act as a transformation function that maps the parameter $\phi$ to the corresponding gate value.

\begin{figure}[h!]
  \centering
  \includegraphics[width=1.0\linewidth]{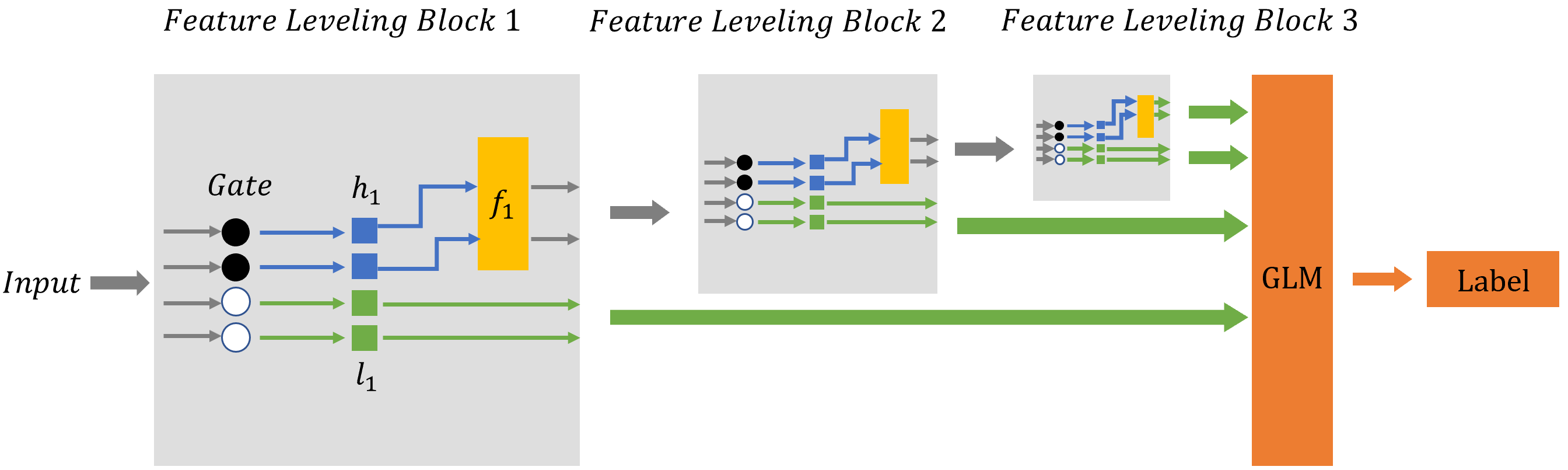}
  \caption{Illustration of the model with three hidden layers. Yellow denotes hidden layer that typically has ReLU activations and green denotes the $k^{th}$ level feature separated out by the gates. Thick arrows denote vector form of input and output. The dimension between the input of the hidden layers and the output can be different.}
\end{figure}

To achieve this discrete gate construction, we propose to learn the gating parameters under the context of $L_0$ regularization. To be able to update parameter values through backpropogation, we propose to use the approximation technique developed by \cite{louizos2017learning} on differentiable $L_0$ regularization. We direct interested readers to the original work for full establishment of approximating $L_0$ and will summarize the key concept in terms of our gating mechanism below.

Although the gate value $z \in \{0, 1\}$ is discrete and the probability for a certain gate to be 0 or 1 is typically treated as a Bernoulli distribution, the probability space can be relaxed by the following: Consider $s$ to be a continuous random variable with distribution $q(s | \phi)$ paramaterized by $\phi$. The gate could be obtained by transformation function $m(\cdot)$ as:
\begin{equation}
    s \sim q(s|\phi), ~ z = m(s) = min(1, max(0,s))
\end{equation}
Then the underlying probability space is continuous because $s$ is continuous and can achieve exactly $0$ gate value. The probability for the gate to be non-zero is calculated by the cumulative distribution function Q:
\begin{equation}
    q(z \neq 0 | \phi) = 1 - Q(s \leq 0 | \phi)
\end{equation}
The authors furthers use the reparameterization trick to create a sampling free noise $\epsilon \sim p(\epsilon)$ to obtain $s$: $s = n( \epsilon, \phi )$ with a differentiable transformation function $n(\cdot)$, and thus $g(\cdot)$ is equivalent to $m\circ n$ where $\circ$ denotes function composition.

Then the objective function under our feature leveling network is:
\begin{equation}
    \begin{aligned}
    &\mathcal{R}(\theta, \phi) = \frac{1}{N}\bigg( \sum_{i=1}^N ( \mathcal{L}(\mathcal{F}(x_i, z;\theta, \phi, B, g), y_i)\bigg) + \frac{\lambda}{K}\sum_{k=1}^K \Big( 1 - Q(s_k \leq 0 | \phi)\Big)\\
    &z_k = g(\phi_k, \epsilon),~~~g(\phi_k, \epsilon) = m \circ n(\phi_k, \epsilon),~~~ \epsilon \sim p(\epsilon)
    \end{aligned}
\end{equation}

\section{Related work}
\textbf{Interpreting existing models: } The ability to explain the reasoning process within a neural network is essential to validate the robustness of the model and to ensure that the network is secure against adversarial attacks \cite{moosavi2016deepfool, brown2017adversarial,gehr2018ai2}. In recent years, Many works have been done to explain the reasoning process of an existing neural network either through extracting the decision boundary \cite{bastani2018verifiable,verma2018programmatically,wang2018efficient,zakrzewski2001verification}, or through a variety of visualization methods \cite{mahendran2015understanding,zeiler2014visualizing,li2015visualizing}. Most of those methods are designed for validation purpose. However, their results cannot be easily used to improve the original models. 

\textbf{Self explaining models} are proposed by \cite{NIPS2018_8003} and it refers to models whose reasoning process is easy to interpret. This class of models does not require a separate validation process. Many works have focused on designing self-explaining architectures that can be trained end-to-end\cite{zhang2018interpretable, Worrall_2017_ICCV,li2018deep,kim2018disentangling, higgins2017beta}. However, most self-explaining models sacrifice certain amount of performance for interpretability. Two noticeable models among these models are able to achieve competitive performance on standard tasks while maintaining interpretability. The NIT framework \cite{NIPS2018_7822} is able to interpret neural decision process by detecting feature interactions in a Generalized Additive Model style. The framework is able to achieve competitive performance but is only able to disentangle up to K groups of interactions and the value K needs to be searched manually during the training process. The SENN framework proposed by \cite{NIPS2018_8003} focuses on abstract concept prototyping. It aggregates abstract concepts with a linear and interpretable model. Compared to our model, SENN requires an additional step to train an autoencoding network to prototype concepts and is not able to disentangle simple concepts from more abstract ones in a per-layer basis. 

\textbf{Sparse neural network training} refers to various methods developed to reduce the number of parameters of a neural model. Many investigations have been done in using $L_2$ or $L_1$ \cite{han2015learning, ng2004feature, wen2016learning,girosi1995regularization} regularization to prune neural network while maintaining differentiability for back propagation. Another choice for regularization and creating sparsity is the $L_0$ regularization. However, due to its discrete nature, it does not support parameter learning through backpropagation. A continuous approximation of $L_0$ is proposed in regard to resolve this problem and has shown effectiveness in pruning both FCNN and Convolutional Neural Networks (CNNs) in an end to end manner \cite{louizos2017learning}. This regularization technique is further applied not only to neural architecture pruning but to feature selections \cite{yamada2018deep}. Our work applies the $L_0$ regularization's feature selection ability in a novel context to select maximum amount of features as direct inputs for the GLM layer. 

\section{Experiments}
We validate our proposed architecture through three commonly used datasets - MNIST, California Housing and CIFAR-10. For each task, we use the same initial architecture to compare our proposed model and FCNN baseline. However, due to the gating effect of our model, some of the neurons in the middle layers are effectively pruned. The architecture we report in this section for our proposed model is the pruned version after training with the gates. The second to last layer of our proposed models is labeled with a star to denote concatenation with all previous $l_k$ and the output of the last hidden layer. For example, in the California Housing architecture, both proposed and FCNN baseline start with $13-64-32-1$ as the initial architecture, but due to gating effect on deeper layers, the layer with $32*$ neurons should have in effect $32 + (13-10) + (64-28) = 71$ neurons accounting for previously gated features. ($13-10=3$ for $l_1$, $64-28 = 36$ for $l_2$).

The two objectives of our experiments are: 1) To test if our model is able to achieve competitive results, under the same initial architecture, compared to FCNN baseline and other recently proposed self-explaining models. This test is conducted by comparing model metrics such as root mean square error (RMSE) for regression tasks, classification accuracy for multi-class datasets and area under ROC curve (AUC) for binary classification. 2) To test if the $k^{th}$ level features gated from the pre-GLM layer make similarly important contributions to the result as features extracted entirely through hidden layers. In order to account for how much each layer's feature contribute to the final decision making, we propose to use the average of absolute values (AAV) of the final GLM layers weights on the features selected by the gates. If the AAV of each level's features is similar, it shows that these features make similar influence on the final decision. 

Experiment implementation details are deferred to supplemental.

\subsection{Datasets \& performances}
\begin{table}
  \caption{MNIST classification and California Housing price prediction}
  \label{sample-table}
  \centering
  \begin{tabular}{lll|lll}
    \toprule
    \multicolumn{3}{c|}{MNIST} & \multicolumn{3}{c}{California Housing}                    \\
    \cmidrule(r){1-3}
    \cmidrule(r){4-6}
    Model        & Architecture      &  Accuracy  & Model        & Architecture      &  RMSE  \\
    \midrule
    FCNN         &  784-300-100-10 &  0.984   &  FCNN         &  13-64-32-1       &  0.529     \\
    L0-FCNN \cite{louizos2017learning}   &  219-214-100-10     &  0.986 & GAM \cite{NIPS2018_7822}          &  -                &  0.506         \\
    SENN (FCNN)         &  784-300-100        &  0.963 & NIT \cite{NIPS2018_7822}          &  8-400-300-200-100-1                &  0.430         \\
    Proposed     &  291-300*-10          &  0.985 & Proposed     &  10-28-32* -1       &  0.477  \\

    \bottomrule
  \end{tabular}
\end{table}
\textbf{The MNIST hand writing dataset} \cite{lecun2010mnist} consists of pictures of hand written digits from 0 to 9 in $28\times 28$ grey scale format. We use a $784-300-100-10$ architecture for both FCNN baseline and the proposed model. This is the same architecture used in the original implementations of \cite{louizos2017learning}. Our model is able to achieve similar result, with less number of layers, as those state-of-the-art architectures using ReLU activated FCNNs . The feature gates completely eliminated message passing to the 100 neuron layer, which implies that our model only need level 1 and level 2 layers for feature extractions to learn the MNIST datasets effectively. 

\textbf{The California Housing dataset} \cite{pace1997sparse} is a regression task that contains various metrics, such as longitude and owners' age to predict the price of a house. 
It contains 8 features and one of the features is nominal. We converted the nominal feature into one-hot encoding and there are 13 features in total. Since California Housing dataset does not contain standard test set, we split the dataset randomly with 4:1 train-test ratio. Our proposed model could beat the FCNN baseline with the same initial architecture. Only 3 out of 13 original features are directly passed to the GLM layer, implying that California Housing's input features are mostly second and third level.

\begin{table}
\caption{CIFAR-10 Binary}
  \label{sample-table}
  \centering
  \begin{tabular}{lll}
    \toprule
    Model        & Architecture      &  AUC  \\
    \midrule
    FCNN         &  3072-2048-1024-2  &  0.855     \\
    GAM  \cite{NIPS2018_7822}   & -       &  0.829 \\
    NIT  \cite{NIPS2018_7822}    &  3072-400-400-1   &  0.860 \\
    SENN (FCNN)         &  3072-2048-1024-2    &  0.856 \\
    Proposed     &  3072-130- 1024*-2  &  0.866 \\
    \bottomrule
  \end{tabular}
\end{table}
\textbf{The CIFAR-10 Dataset} \cite{krizhevsky2014cifar} consists of $32\times 32$ RGB images of 10 different classes. We test our model's ability to extract abstract concepts. For comparison, we follow the experiments in the NIT paper and choose the class cat and deer to perform binary classification. The resulting architecture shows that for FCNN networks, most of the the two chosen classes are mainly differentiated through their second level features. None of the raw inputs are used for direct classification. This corresponds to the assumption that RGB images of animals are relatively high level features. 

\subsection{$k^{th}$ level feature passage and AAV of GLM weights}
We also validate the percentage of input to the $k^{th}$ hidden layer, which is the $k^{th}$ level features selected by the gates. We also measures to what extent could these features contribute to the final decision(Figure 4). Through inspecting the percentage of features that flow to the GLM layer (the total number of gate with 1 as its value) and the AAV metric that we mentioned in the prior section, we notice that $k^{th}$ level features generally have similar, if not higher, AAV weights compared to the features extracted through all hidden layers. This implies that the $k^{th}$ level features are making similar contribution to the decision as those features extracted by FCNNs alone.

\section{Strength in pruning redundant hidden layers}
Due to our proposed model's ability to encourage linearity, our model is also able to reduce its network complexity automatically by decreasing the number of hidden layers. Empirically, as training goes on, each layer observes increasing number of features flowing to the GLM. Thus, more $l_k$ features are transported directly to the GLM, reducing complexity of our model. This implies that our network is learning to use more $l_k$ features directly in GLM as opposed to transforming these features in further hidden layers. 

 We also observe that for some tasks such as MNIST classification, when the dataset feature level is less than the number of hidden layers, our proposed model can learn to prune excess hidden layers automatically as the network learns not to pass information to further hidden layers. As a result, the number of hidden layers are effectively reduced. Therefore, we believe that our framework is helpful for architectural design by helping researchers to probe the ideal number of hidden layers to use as well as understanding the complexity of a given task.

\begin{figure}[h!]
\centering
\begin{minipage}{.45\textwidth}
\centering
\includegraphics[width=1\linewidth]{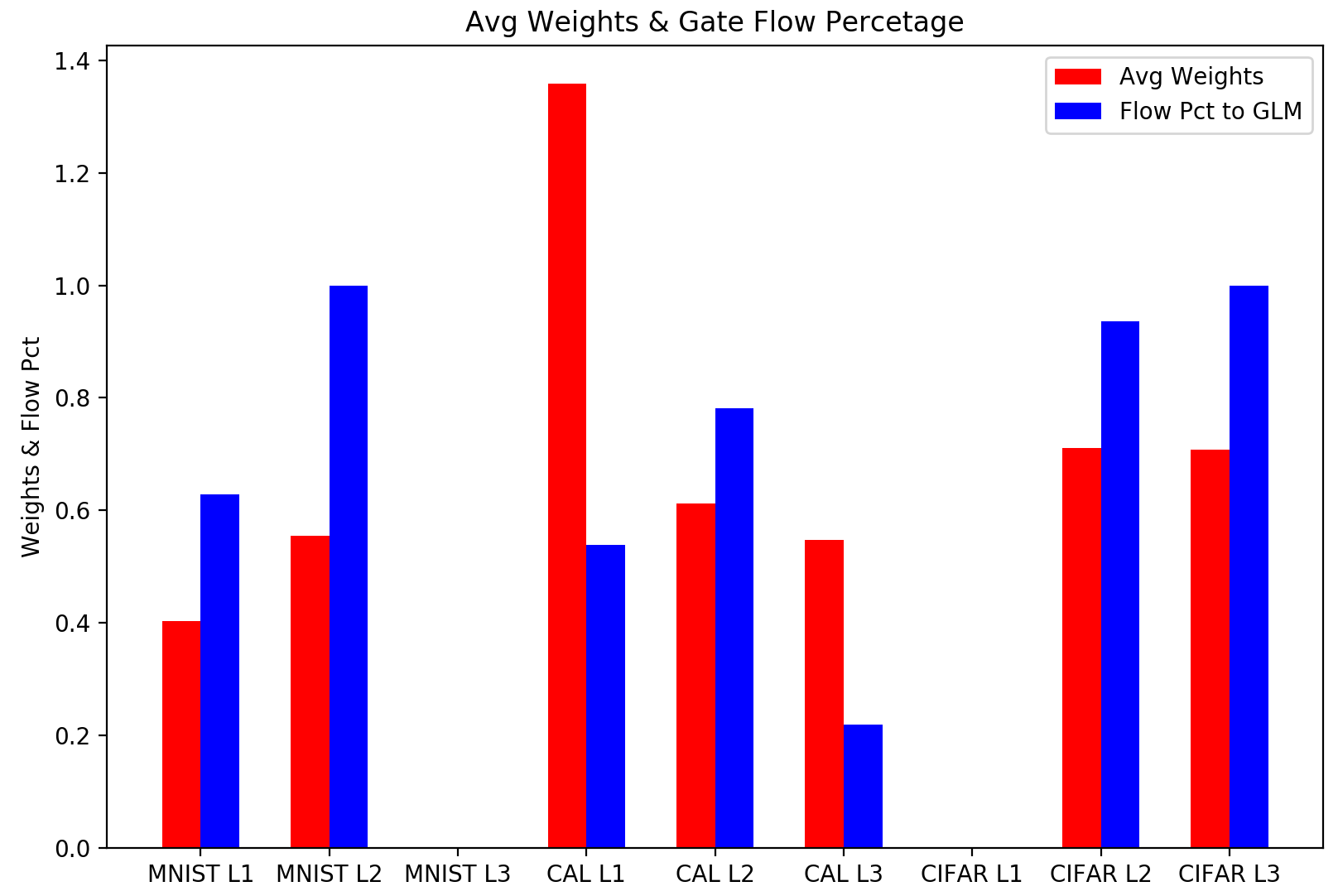}
\end{minipage}\qquad
\begin{minipage}{.45\textwidth}
\centering
\includegraphics[width=1\linewidth]{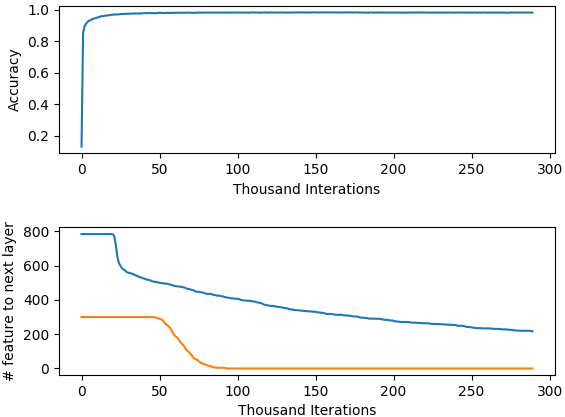}
\end{minipage}

\bigskip

\begin{minipage}[t]{.45\textwidth}
\centering

\caption{The percentage of gated features and average absolute weight (AAV) in GLM at different levels for all test models. Cal-Housing's AAVs are scaled down for graphing clarity. }
\label{camellia}
\end{minipage}\qquad
\begin{minipage}[t]{.45\textwidth}
\centering
\caption{MNIST training performance curve and number of inputs passed to the following hidden layer (blue denotes the number of features passed to the firs hidden layer. Orange curve denotes the second).}
\label{rose}
\end{minipage}
\end{figure}

\section{Discussion}
In this work we propose a novel architecture that could perform feature leveling automatically to boost interpretability. We use a toy example to demonstrate the fact that not all features are equal in complexity and most DNNs take mixed levels of features as input, decreasing interpretability. We then characterize absolute feature complexity by the number of layers it requires to be extracted to make GLM decision. To boost interpretability by isolating the $k^{th}$ level features. We propose feature leveling network with a gating mechanics and an end-to-end training process that allow the $k^{th}$ level features to be directly passed to the GLM layer. We perfrom various experiments to show that our feature leveling network is able to successfully separate out the $k^{th}$ level features without compromising performance.

There are two major directions for extension based on our proposed architecture: The first one is to extend our current construction to the context of convolutional neural networks. Another direction is to associate our network's identity mapping of low level features with residual operations such as ResNet \cite{he2016deep}, Highway  Network \cite{srivastava2015training} and Dense Network \cite{huang2017densely} and try to gain insights into their success. \cite{hardt2016identity}. 

\bibliography{macro}
\bibliographystyle{plainnat}

\section{Revisit GLM for interpretations of deep neural networks}
Consider training a linear model with dataset $\{\mathcal{X}, \mathcal{Y}\}$ where $\mathcal{X}$ is the set of features and $\mathcal{Y}$ is the corresponding set of labels. The goal is to learn a function $f(x)$ from $(x_i, y_i) \in \{\mathcal{X}, \mathcal{Y}\}$subject to a criteria function $\mathcal{L}_{\theta}(x_i, y_i)$ with parameter set $\theta$.

In a classical setting of Linear Models, $\theta$ usually refers to a matrix $w$ such that:
\begin{equation}
    \hat{y} = f(x) = T( w^{\top}x + \beta )
\end{equation}
Here, $\hat{y}$ refers to the predicted label given a sample instance of a set of feature $x$ and T refers to the set of functions such as Logictic, Softmax and Identity. GLM is easy to interpret because the contribution of each individual dimension of x to the decision output y by its corresponding weight. Therefore, we hope to emulate GML's interpretability in a DNN setting - by creating a method to efficiently back-trace the contribution of different features. 

We argue that our proposed architecture is similar to a GLM in that our final layer makes decision based on the weights assigned to each level of input features. Our model is linear in relationship to various levels of features. Given k levels of features, our model makes decision with $y = [w_1^{\top}l_1,w_2^{\top}l_2, ..., w_K^{\top}l_K]$, each weight parameter $w_i$ indicates the influence of that layer. With this construction, we can easily interpret how each levels of feature contribute to decision making. This insight can help us to understand whether the given task is more "low level" or "high level" and thus can also help us to understand the complexity of a given task with precise characterization. 

\subsection{The last layer of common neural networks is a GLM layer }
The "classical" DNN architecture consists of a set of hidden layers with non-linear activations and a final layer that aggregates the result through sigmoid, softmax, or a linear function. The final layer is in fact similar to the GLM layer since it itself has the same form and optimization objective.

\section{Reproducing empirical results}
\subsection{General configuration}
All models are implemented in TensorFlow\cite{abadi2016tensorflow} and hyperparameters configurations could be found in our public repository or supplemental code. Model name with citation denotes that the result is obtained from the original paper. SEEN's architecture listed is the prototyping network while we use similar architecture for autoencoder parts. All SENN models are re-implemented with fully connected networks for comparison purposes.

\subsection{Dataset and preprocessing}
MNIST is a dataset that contains 60000 training and 10000 testing of handwriting digits from 0 to 9. Experiment results were tested against the allocated testing set.

CIFAR-10 is a dataset consists of 10 classes of images each with 10000 training and 2000 testing. We used the allocated testing set for reporting results.

For MNIST, CIFAR-10, we rescaled the color channel with a divisor of 255., to make pixel values from 0 to 1.

For Cal Housing, we dropped all samples with any empty value entry. Normalize all numerical values with mean and standard deviation. 

The IXOR dataset is generated with the script attached in the supplemental material under src/independent$\_$xor.

\subsection{Hyperparameter}
The only tunable hyperparameter in our model is the $\lambda$ which we usually consider values from 0.5 to 0.01. All the $\lambda$ values to display result is in the model scripts of the attached folder. Generally, lower $\lambda$ are better for training more complicated dataset such as CIFAR-10 to prevent too many Gating at early stage. 

\subsection{Exact number of iteration runs}
MNIST 280000 \\
CIFAR-10 680000 \\
California Housing 988000 \\
\end{document}